# Arabic Inquiry-Answer Dialogue Acts Annotation Schema

AbdelRahim A. Elmadany[1], Sherif M. Abdou[2], Mervat Gheith[1]

[1]*Institute of Statistical Studies and Research (ISSR), Cairo University*
*Emails: ar_elmadany@hotmail.com, mervat_gheith@yahoo.com*
[2]*Faculty of Computers and Information, Cairo University*
*Email: sh.ma.abdou@gmail.com*

***Abstract:*** *We present an annotation schema as part of an effort to create a manually annotated corpus for Arabic dialogue language understanding including spoken dialogue and written 'chat' dialogue for inquiry-answer domain. The proposed schema handles mainly the request and response acts that occurs frequently in inquiry-answer debate conversations expressing request services, suggests, and offers. We applied the proposed schema on 83 Arabic inquiry-answer dialogues.*

## I.    Introduction

Arabic Natural Language Processing (ANLP) including Arabic Language Understanding (ALU) within dialogue-based research has gained an increasing interest in the last few years. Building an ALU system requires an annotated dialogue acts (DAs) corpora that is annotated according to a specified dialogue acts schema.

Inquiry-answer is an important part of industry world because it is the core of a successful business. When quality services are met and exceed customer expectations, the customers are pleased. In order to ensure this happens, service managers (inquiry-answer operators) need to be aware of the parts of the service delivery experience that are open to cultural effects[1].

In this paper, we propose a dialogue acts annotation schema for Arabic inquiry-answer dialogues. This schema defines the dialogue acts using multi-dimensions within communication functions based on request and response that have certain questions, responses and the relation between them.

In this paper, we proposed the first version of Arabic inquiry-answer manually annotated dialogues corpus, which contains 83 spoken and written 'chat' dialogues.

In the following sections, section 2 summarizes the main annotation schemas developed for English language and Arabic language. Section 3 describes the proposed annotation schema. Section 4 describes the annotation corpus and section 5 includes the conclusion.

## II.    Related works

Dialogue act refers to the meaning of an utterance representation at the level of illocutionary force [2]. Speech acts terminology has been addressed by Searle (1969) based on Austin (1962) work.

The idea of a dialogue act plays a key role in studies of dialogue, especially in communicative behavior understanding of dialogue participants, in building annotated dialogue corpora and in the design of dialogue management systems for spoken human-computer dialogue.

Dialogue act is approximately the equivalent of the speech act of Searle (1969). Dialog acts are different in different dialog systems. The research on dialog acts has increased since 1999, after spoken dialog systems became a commercial reality[3].

The MapTask project [4] proposed labeling schema using 12 dialogue acts based on two categories (1) initiating moves (2) response. The VERBMOBIL project (1993-2000) aimed at the development of an automatic speech to speech translation system for the languages German, American English and Japanese [5]. The VERBMOBIL Project had two phases, the first phase proposed labeling schema using hierarchy of 43 dialogue acts [6]. The second phase expanded the dialogues from meeting scheduling to comprehensive travel planning. Thus change labeling schema to a hierarchy of 18 dialogue acts [7].

The DAMSL Dialogue Act Markup using Several Layers (DAMSL) were proposed as a general-purpose schema [8-10] developed for multi-dimensional dialogue acts annotation. SWITCHBOARD-DAMSL is improved version of DAMSL proposed by [11]when they wanted to annotate a large amount of transcribed speech data 'Switchboard Corpus' because of the difficulty of consistently applying the DAMSL annotation schema [11, 12]. SWITCHBOARD-DAMSL schema includes 220 dialogues acts, but it is still difficult to be used for manual annotation because it is a very large set. Moreover, [11]reported 0.80 of Kappa score with the 220 dialogue acts and 130 dialogue acts occurred less than 10 times in the entire corpus [13]. To obtain enough data per class for statistical modeling purposes, [11]proposed new dialogue act schema namely SWITCHBOARD contains 42 mutually exclusive dialogue acts types.

The ICSI-MRDA Meeting Room corpus use DA variant of the DAMSL dialogue acts schema like the SWITCHBOARD corpus by combining the tags into single, distinct dialogue acts to reduce aspects of the multidimensional nature of the original DAMSL annotation scheme. There are 11 general tags and 39 specific acts that are used over ICSI-MRDA Meeting Room corpus [14]. The AMI project, a European research project centered on multi-modal meeting room technology, uses 15 dialogue acts.

Dynamic Interpretation Theory (DIT) [15] reported dialogue acts schema with a number of dialogue act types from DAMSL [8] and other schema. The DIT++ is a comprehensive system of dialogue act types obtained by extending the acts of DIT [16]. DIT++ schema has 11 dimensions with around 95 communicative functions, around 42 of which, like switchboard are for general purpose functions, whereas others cover elements of feedback, interaction management and the control of social obligations [13].

[17]has proposed a preliminary version of ISO DIS 24617-2:2010 as an international standard for annotating dialogue with semantic information; in particular concerning the communicative functions of the utterances, the kind of content they address, and the dependency relations to what was said and done earlier in the dialogue. [18]has proposed the final version of ISO Standard 24617-2.

Most of the previous proposed annotation schemes to mark-up dialogue corpora in languages such as English, German and Spanish. There are few efforts were done to propose dialogue acts annotation schema for Arabic. First attempt was by [19] which proposed a DAs classifier with scheme that contains about 10 DAs (assertion, declaration, denial, expressive evaluation, greeting, indirect request, question, promise/denial, response to question, and short-response).

(Dbabis *et al.*, 2012) proposed another DAs schema within an Arabic discussion that contains about 13 DAs within 6 categories: Social Obligation Management (e.g. Opening, Closing, Greeting … etc.), Turn Management (e.g. Acknowledgement, Calm, Clarify Request … etc.), Request (e.g. Question, Order, Promise … etc.), Argumentation (e.g. Opinion, Appreciation, Accept, Reject … etc.), Answer, and Statement [20].

These schemas were used to mark-up dialogue corpora based on a general conversion discussion like TV talk-show programs.

There are two other works concerned with understanding the Arabic inquiry-answer dialogue conversions, [21] proposed an annotation schema to semantically label words on spoken Tunisian dialect turns which are not segmented into utterances as shown in Figure 1. This schema was applied on TUDICOI corpus that contains dialogues collected from the National Company of Railway in Tunisia (SNCFT). This schema was used to mark-up word-by-word in dialogue turn not turns or utterances dialogue acts. This schema contains about 33 semantic labels for word annotation within 5 dimensions e.g. Train_Type, Trip_Time, and Destination.

[ مع [Out] ] وقتاش [Hour_Req] ] إي[Out] ] إي[Out] ] التران[Train] ] يمشي Departure_Cpt]

*Figure 1 Example of semantic labeling from [21]*

[22] proposed a dialogue acts classifier based on hotel reservations corpus that consist of 100 queries of different types (negation, affirmation, interrogation and acceptance). They used a small test set that contained 140 utterances with 14 different dialogue acts that is uttered spontaneously to evaluate their system. This dialogue acts are incomplete to build a inquiry-answer discourse structure because this tag set can't annotate communicative functions related to social obligations, suggestions, miss understanding correction, warnings, explanation … etc.

### III.     Annotation Schema

In this work, we started from dialogue acts in a previous annotation schema for either Arabic or English. We proposed a general dialogue acts annotation schema for inquiry-answer based request and response dimensions. Table 1,Table 2, and Table 3 describe the proposed dialogue acts based on request dimension within 6 dialogue acts, response dimension within 14 dialogue acts, and other dimension within 3 dialogue acts.

| **Request Acts** |
| --- |
| Dealing with request activities and tasks |
| Taking-Request<br>*Dealing with taking request e.g. hello* |
| Service-Question<br>*Dealing with services request e.g. asking about service information or required a service.* |
| Confirm-Question<br>*Happens when needs to confirmation about some information.* |
| YesNo-Question<br>*Happens when needs Yes or No answer.* |

| Choice-Question |
| --- |
| *Happens when needs select one answer from service multiple-choices question.* |
| Other-Question |
| *Happens when asking about non-service question e.g. mobile number, email, or address.* |
| Turn-Assign |
| *Happens when wants to addressee the speaker to take the turn e.g. Adam?* |

*Table 1: Request dimension dialogue acts*

| **Response Acts** |
| --- |
| Dealing with response activities and tasks |
| Service-Answer |
| *Happens when answer a Service-Question or Choice-Question.* |
| Other-Answer |
| *Happens when answer an Other-Question.* |
| Agree |
| *Describe agreement/accept answer from Confirm-Question or YesNo-Question.* |
| Disagree |
| *Describe disagreement/reject answer from Confirm-Question or YesNo-Question.* |
| Greeting |
| *Happens when speaker wants to greeting and welcome the other speaker. Also describe greeting accept 'return-greeting'.* |
| Inform |
| *Happens when speaker wants to explain or describe something to other speaker.* |
| Thanking |
| *Happens when speaker wants to thank the other speaker. Also describe thanking accept.* |
| Apology |
| *Happens when speaker wants to apology.* |
| MissUnderstandingSign |
| *Happens when non-understanding the previous utterance.* |
| Correct |
| *Happens when correct an information in previous utterance or in current utterance.* |
| Pausing |
| *Happens when needs to request more time or stealing time e.g. just a moment.* |
| Suggest |
| *Happens when provides a suggestion.* |
| Promise |
| *Happens when provides a promise.* |
| Warning |
| *Happens when provides a warning action.* |
| Offer |
| *Happens when provides an offer to the customer.* |

*Table 2: Response dimension dialogue acts*

| **Other Acts** |
| --- |
| Dealing with neither request nor response activities and tasks |
| Opening |
| *Dealing with opening obligation utterance e.g. "Good evening, Banque Misr, Ahmed Samy speaking".* |
| Closing |
| *Dealing with closing obligation request e.g. "Thank you for calling and goodbye".* |
| Self-Introduce |
| *Happens when wants to introduce our self or organization.* |

*Table 3: Other dimension dialogue acts*

We set 'Agree' dialogue act to describe the agreement/acceptance answers as accept-confirmation, Yes-Answer, accept-thanking, and accept-apology. In contrast, we set 'Disagree' dialogue act to describe the disa-greement/reject answer as disconfirm, No-Answer, and Reject-thanking, Reject-apology. Our empirical analysis for our corpus leaded to identifying 'Opening' dialogue act as a group of 'Greeting' act and 'self-Introduce' act. For example from our corpus:

| Arabic: | مساء الخير بنك مصر احمد مع حضرتك |
|---|---|
| *Buckwalter:* | *msA' Alxyr bnk mSr AHmd mE HDrtk* |
| English: | Good evening, Banque Misr, Ahmed speaking |

This turn tagged as 'Opening' dialogue act and segmented to three utterances as:

- مساء الخير/*msA' Alxyr/Good evening* is tagged as 'Greeting' dialogue act.
- بنك مصر /*bnk mSr/Banque Misr* is tagged as 'Self-Introduce' dialogue act.
- احمد مع حضرتك /*AHmd mE HDrtk/Ahmed speaking* is tagged as 'Self-Introduce' dialogue act.

## IV. Annotation Corpus

For the best of our knowledge, there is no corpus for annotated Arabic spoken or written 'chat' dialogues in Arabic dialect. For this reason, we built our own corpus as the first version of manually annotated corpus for Arabic dialogue language understanding including spoken dialogue and written 'chat' dialogue for inquiry-answer domain.

*Figure 2: Utterance Annotation Example*

This corpus contains two parts (1) spoken dialogue which contains 52 phone calls recorded from Egyptian's banks inquiry-answer and Egypt Air Company with an average duration of two hour of talking time after removing ads from calls. It consists of human-human discussions about providing services e.g. create new bank account, service request, balance check and flight reservation. These phone calls were transcribed using Transcriber®, a tool that is frequently used for segmenting, labeling and transcribing speech corpora. (2) Written 'Chat' dialogues, which contain 30 chat dialogues, recorded from mobile agencies web sites 'KSA Zain, KSA Mobily, and KSA STC'.

This corpus contains 3001 turns with average 6.7 words per turn and contains 4727 utterances with average 4.3 words per utterance. This corpus is manually segmented and annotated using the proposed schema. An annotation tool was developed to manually annotate dialogue turns and segment it is to utterances and label the dialogue act. A sample-annotated turn is shown in Figure 2.

## V. Conclusion

In this paper, we proposed a general dialogue acts schema for Arabic inquiry-answer dialogue either spoken or written 'chat'. We proposed the first version of the manually annotated inquiry-answer dialogue as a part of Arabic language understanding methodologies.

In addition, we introduced the first version of annotated corpus that includes 83 inquiry-answer dialogues from banks, Airlines Company, and mobile networks agency. In future work we plan to enrich this corpus with inquiry-answer dialogues from other domains e.g. Online Markets, and Railway Networks in both forms spoken and written 'chat' dialogues.